\documentclass[9pt,twocolumn,twoside]{opticajnl}
\journal{opticajournal} 
\setboolean{shortarticle}{true}
\usepackage{soul}
\usepackage{lineno}
\usepackage[normalem]{ulem}

\title{3D Imaging of Complex Specular Surfaces by Fusing Polarimetric and Deflectometric Information }

\author[1,2,*]{Jiazhang Wang}
\author[2]{Oliver Cossairt}
\author[1,2]{Florian Willomitzer}

\affil[1]{Wyant College of Optical Sciences, University of Arizona, Tucson, AZ 85721}
\affil[2]{Department of Electrical and Computer Engineering, Northwestern University, Evanston, IL 60208}

\affil[*]{jiazhangwang@arizona.edu}

\begin{abstract}
 Accurate and fast 3D imaging of specular surfaces still poses major challenges for state-of-the-art optical measurement principles. Frequently used methods, such as phase-measuring deflectometry (PMD) or shape-from-polarization (SfP), rely on strong assumptions about the measured objects, limiting their generalizability in broader application areas like medical imaging, industrial inspection, virtual reality, or cultural heritage analysis.
In this paper, we introduce a measurement principle that utilizes a novel technique to effectively encode and decode the information contained in a light field reflected off a specular surface. We combine polarization cues from SfP with geometric information obtained from PMD to resolve all arising ambiguities in the 3D measurement. Moreover, our approach removes the unrealistic orthographic imaging assumption for SfP, which significantly improves the respective results. We showcase our new technique by demonstrating single-shot and multi-shot measurements on complex-shaped specular surfaces, displaying an evaluated accuracy of surface normals below $0.6^\circ$.

\end{abstract}

\setboolean{displaycopyright}{false} 

\begin{document}

\maketitle

\begin{figure*}[t]
\centering
\includegraphics[width=\linewidth]{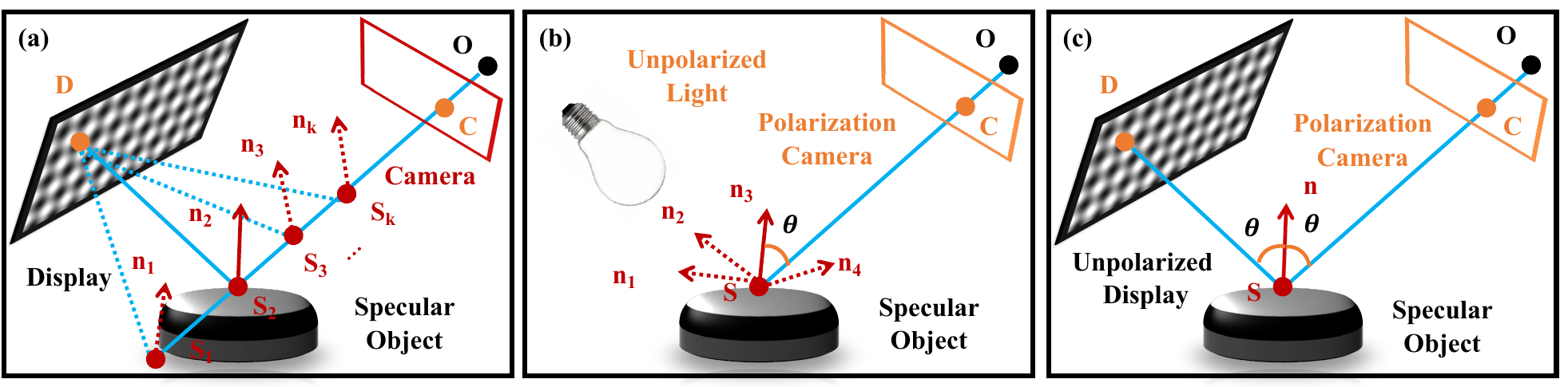}
\caption{\textbf{3D imaging of specular surfaces: Shortcomings of current methods and our solution:} a) Normal-depth ambiguity in phase-measuring deflectometry (PMD): Without knowing the position $S$ of each object surface point along the camera ray, the respective surface normal can not be retrieved. b) In shape from polarization (SfP), the ambiguity in two potential zenith angles and two potential azimuth angles leads to four candidates for each surface normal. Moreover, the unrealistic orthographic assumption leads to high normal errors in off-center image regions. c) Our proposed solution combines geometric information from PMD and polarization information from SfP, to calculate the shape and normal map of the specular object surface in single-shot with high accuracy, free from any ambiguity.}
\label{fig1}
\end{figure*}

\section{Introduction}

Accurate, robust, and fast 3D reconstruction of specular surfaces plays a crucial role in industrial inspection \cite{olesch2014deflectometric,hofer2016infrared}, cultural heritage preservation and analysis \cite{gomes20143d,willomitzer2020hand}  or medical imaging~\cite{liang2016single}, with recent potential implementations in novel viewpoint rendering approaches \cite{mildenhall2021nerf,guo2022nerfren},  which can be specifically tailored to specular objects \cite{tiwary2023orca}. However, state-of-the-art 3D imaging methods still face significant challenges in measuring specular surfaces, especially complex-shaped objects. Due to the large potential impact of a robust solution, the problem of 3D reconstruction of specular surfaces has been tackled by several scientific communities in the past, in particular the optical metrology and computer vision community.

\noindent \textbf{Phase Measuring Deflectometry (PMD)} \cite{knauer2004phase, faber2012deflectometry, liang2016single, willomitzer2020hand} is frequently used in the optical metrology community and stands out for its ability to achieve up to sub-micron depth resolution using relatively inexpensive off-the-shelf components (display and camera). Besides its wide use in industrial inspection of optical components such as lenses or mirrors  \cite{knauer2004phase,graves2018model}, PMD has recently been employed in diverse fields, including medical diagnosis~\cite{liang2016single}, mobile measurement \cite{willomitzer2020hand}, eye-tracking \cite{wang2023accurate, wang2023optimization,choi2024accurate}, and cultural heritage preservation \cite{willomitzer2020hand}.  In PMD, the reflection of a pattern displayed on a screen is observed with a camera over the specular surface (see Fig. \ref{fig1}(a)). From the resulting deformation of the pattern in the camera image, the surface normals, and later the shape can be calculated. Standard PMD suffers from height-normal ambiguities \cite{knauer2004phase, huang2018review} that arise when estimating the surface normal with a single camera and display, as the incident ray from the display cannot be determined solely based on its corresponding emission pixel (see Fig. \ref{fig1}(a)). For this reason, many PMD approaches rely on strong prior assumptions about the object surface (such as its position in space or approximate shape) to obtain the normal fields \cite{huang2011dynamic}. 
Other common approaches mitigate this issue by using a second camera or a second display to provide missing additional information \cite{knauer2004phase, huang2018review}. However, such solutions increase setup complexity and introduce additional calibration challenges. 

\noindent \textbf{Shape-from-Polarization (SfP)} \cite{rahmann2001reconstruction,atkinson2006recovery,morel2005polarization,kadambi2015polarized} is a 3D measurement principle that gained significant popularity in the computer vision community and has developed into a well-established qualitative 3D imaging method, with recent integrations in deep learning-based \cite{ba2020deep} and event-based \cite{muglikar2023event} techniques. In this paper, we focus on the ability of SfP to measure \textit{specular surfaces}, although it has been demonstrated for diffuse surfaces as well. Typical SfP setups illuminate the object surface with an unpolarized light source, such as a thermal light bulb or the sun. As described by the Fresnel equation and Snell's law, the unpolarized incident light becomes partially polarized upon reflection or scattering off the object surface~\cite{atkinson2006recovery}. Eventually, the reflected or scattered light is captured with a polarization-sensitive imager, such as a camera with a linear polarizer that is rotated in front, or a polarization camera \cite{lucidpolsensor}, which has a grid of polarizers on top of the pixel grid and allows for polarization measurements in single-shot. As the change in polarization depends on the slope of the measured surface w.r.t. light source and camera position, the normal map of the surface can be extracted. This process, however, is also not without severe limitations:  The method assumes an orthographic ray model \cite{rahmann2001reconstruction,atkinson2006recovery}, meaning that all camera rays of the camera observing the object are assumed to be parallel. As this condition can typically not be satisfied without sophisticated optical setups (which would introduce additional limitations), the orthographic assumption typically produces severe normal vector errors in the off-center image regions ($\sim 5^\circ$ up to even $25^\circ$ (!), see below and \cite{atkinson2006recovery}).  Another problem of SfP is additional ambiguities in the evaluated surface normal vector field (see Fig. \ref{fig1}(b)), which requires prior information about the measured surface or additional information from a second sensor for a solution. In this regard, the measurement of specular surfaces with SfP is particularly challenging, as these surfaces lead to more ambiguities (azimuth and elevation, see below), compared to diffuse surfaces (azimuth only).
To resolve the normal ambiguities, prior work has combined SfP with other imaging principles, such as shape from shading~\cite{smith2016linear}, multispectral measurements \cite{huynh2013shape}, or coarse depth map measurements from a secondary depth sensor \cite{kadambi2015polarized, cui2017polarimetric}. However, those methods are mainly limited to diffuse surfaces and still rely on the orthographic assumption. A robust SfP solution for specular surfaces does not exist to our knowledge. \\

\noindent In this paper, we introduce a novel method for the absolute shape and normal measurement of specular surfaces. Our method fuses geometric information from deflectometry measurements together with polarization cues to recover an accurate reconstruction of the normal map and shape of the measured surface. Due to the unique combination of different information sources, our method is able to analytically resolve all normal and depth ambiguities in the system. Moreover, our approach does not rely on the unrealistic orthographic assumption, which significantly improves the accuracy of the captured normal fields compared to conventional SfP.

\section{Mathematical framework}

\label{principle}

\label{Def} \noindent \textbf{Ambiguities in Deflectometry:} In a PMD setup, a display illuminates the object with a known pattern (e.g., fringe pattern, see Fig. \ref{fig1}(a)). The pattern is reflected by the measured surface and imaged onto the camera chip. For each display point $D(x_d,y_d,z_d)$, the corresponding point on the camera chip $C(x_c,y_c,z_c)$ can be found, e.g., via  phase-shifting methods \cite{knauer2004phase, huang2018review, su2010software}, Fourier transform based methods \cite{takeda1983fourier, Kemao:04}, or via deep learning \cite{feng2019fringe}.
Together with the known camera center $O$, the camera ray vector  $\overrightarrow{OC}$ can be defined for each ray originating from each camera pixel. As shown in Fig. \ref{fig1}(a), this ray passes through its corresponding point \(S(x_s,y_s,z_s)\) on the object surface. Eventually, $S$ is connected with $D$ to define the incident ray vector $\overrightarrow{SD}$. The bisector vector between the reflected ray vector $\overrightarrow{SC}$ and the incident ray vector $\overrightarrow{SD}$ represents the surface normal $\vec{n}$. In other words, $\vec{n}$ is a function of $D$, $C$, $O$, and the location of surface point $S$ on the camera ray $\overrightarrow{OC}$:
\begin{equation}\label{eq:s}
    \overrightarrow{OS}=t\cdot \overrightarrow{OC}, ~~ t>1
\end{equation}
\begin{equation}
    \vec{n} = f(D,C,O,t)
    \label{eq:pmd}
\end{equation}

\noindent In conventional PMD, the standoff distance of the surface to the camera (which is defined by $t$) is unknown. Different values for $t$ would lead to different stand-off distances of the surface point $S$, which results in different surface normals $\vec{n}$ for the same $D$, $C$, $O$ (see Fig. \ref{fig1}(a)). This means that stand-off distance and surface normal can not be retrieved at the same time. The problem is known as the "normal-depth-ambiguity" of PMD. \\

\label{sfp} \noindent \textbf{Ambiguities and orthographic projection assumption in Shape-from-Polarization:} In SfP, the recorded intensity can be represented as a sinusoidal function of the angle of the polarizer $\phi_{\text{pol}}$:
\begin{equation}
    I(\phi_{pol}) = \frac{I_{max}+I_{min}}{2}+\frac{I_{max}-I_{min}}{2} \cdot cos(2(\phi_{pol}-\varphi))
    \label{eq:sin}
\end{equation}
$\varphi$ denotes the polarization angle of the light reflected off the object surface. The maximum and minimum intensities $I_{max}$ and $I_{min}$ occur if this light is parallel or perpendicularly polarized w.r.t. the polarizer.  $I_{max}$ and $I_{min}$ are utilized to define the degree of polarization (DoP)
\begin{equation}
    \rho = \frac{I_{max}-I_{min}}{I_{max}+I_{min}} ~~,
    \label{eq:DoP_I}
\end{equation}
which can be also written as a function of the incident angle $\theta$ of the light ray w.r.t. the surface normal vector:
\begin{equation}\label{eq:dop}
    \rho = \frac{2sin^2\theta cos\theta \sqrt{m^2-sin^2\theta}}{m^2-sin^2\theta-m^2sin^2\theta+2sin^4\theta}
\end{equation}
 \(m\) denotes the refractive index of the material. For non-dielectric surfaces, the  DoP of \eqref{eq:dop} can be approximated \cite{morel2005polarization} by 
 
\begin{equation}
    \rho_{non-dielectic} = \frac{2m~tan\theta~sin\theta}{tan^2\theta~sin^2\theta+m^2(1+\kappa^2)}~~,
    \label{eq:dop_metal}
\end{equation}
where $\kappa$ is the material's attenuation. As discussed, state-of-the-art SfP approaches assume an orthographic ray model. This model makes the simplification that each ray reflected back from the object surface into the camera is perfectly perpendicular to the image plane. \textit{Only under this specific assumption}, the incident angle $\theta$ (which is also the angle of reflection $\theta$) is equal to the zenith angle of the surface normal (spherical camera coordinates). The azimuth angle $\alpha$ is defined as the angle between the projection of the surface normal on the x-y-image plane and the x-axis of the image plane. Therefore, standard SfP estimates the surface normal with the orthographic assumption via:
\begin{equation}
    \Vec{n} = (sin\theta cos\alpha, sin\theta sin\alpha, cos\theta)
\end{equation}

\noindent Estimating the surface normal of specular (e.g., metallic) surfaces with this procedure leads to several problems:  First, a DoP measurement via \eqref{eq:DoP_I} and  \eqref{eq:dop_metal} leads to multiple candidates for surface normals, as one value of $\rho$ always corresponds to two potential incident angles $\theta$. The azimuth angle shows ambiguity as well: $\alpha=\varphi \pm \pi/2$ for specular reflections. All these ambiguities sum up to four possible normals for each surface point (see Fig. \ref{fig1}(b)).  In addition to this normal ambiguity,  the orthographic model assumption in SfP is too crude for common perspective cameras and leads to severe errors in surface normal estimation of $\sim 5^\circ$ up to even $25^\circ$ (!) \cite{atkinson2006recovery,kadambi2015polarized} depending on the angular field of view with further error propagation in the shape estimation from the captured normals. This might be the reason why  SfP is commonly not used for quantitative high-accuracy surface measurements (e.g., in optical metrology).

Therefore, a polarization-based approach capable of avoiding the orthographic assumption and resolving all possible normal ambiguities is of great interest for high-quality quantitative surface measurements. It has the potential to bridge the fields of computer vision and optical metrology and usher in a new wave of application opportunities at the intersection of both fields.

\section{Our solution}

\begin{figure*}[h]
\centering
\includegraphics[width=\linewidth]{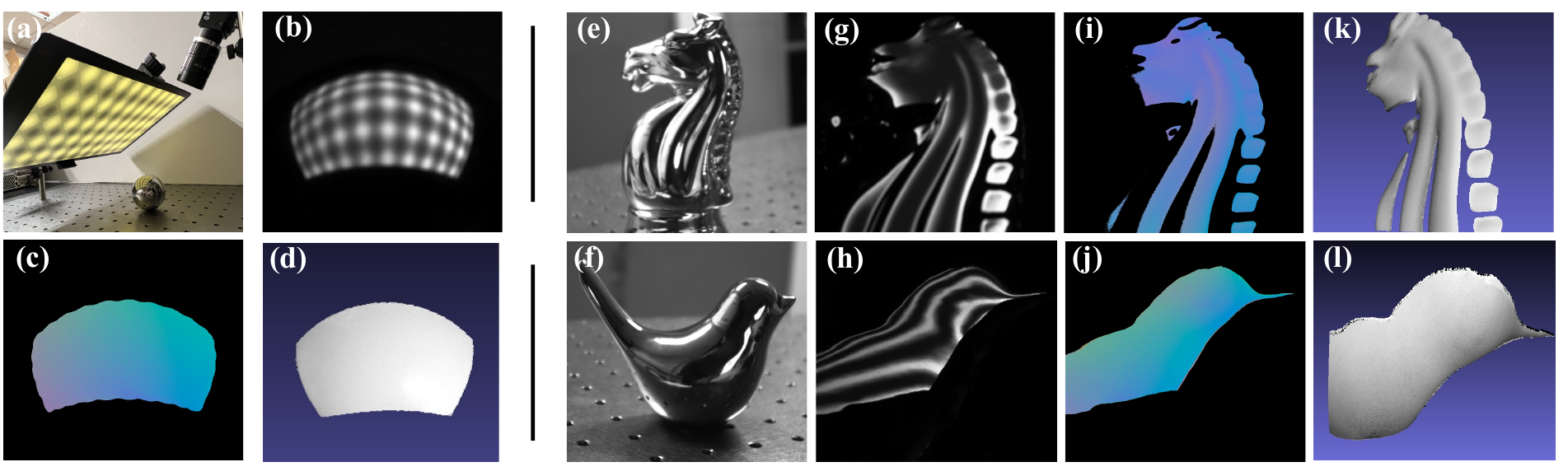}
\caption{\textbf{Quantitative and Qualitative measurements.} (a) Experimental setup: Quantitative measurement of a bearing ball with known size. (b) Sample polarization camera image. (c) Retrieved normal map (RMSE normal error: $0.6^\circ$). (d) 3D surface shape. (e)-(l) Qualitative measurements of two objects (horse and bird) with complex shapes (high surface frequencies). Columns from left to right: object pictures, captured sample camera images, evaluated surface normal map, evaluated 3D shape.}
\label{results}
\end{figure*}

\textbf{Approach:} Our novel 3D imaging approach utilizes polarization information in conjunction with deflectometric information for the fast and accurate 3D measurement of specular surfaces. By jointly addressing the discussed shortcomings of PMD and SfP, we demonstrate simultaneous absolute surface normal and surface shape measurements in a multi-shot and even single-shot fashion \cite{Wang2024Polar}. Our setup (see Fig. \ref{fig1}(c) and Fig. \ref{results}(a) consists of an unpolarized display (e.g., e-ink display) which displays a known pattern, and a polarization camera able to capture four images at four different polarizer angles in single-shot. After image acquisition, the display-camera correspondence is evaluated from one of the four images as described in Sec.\ref{principle}. As discussed, the correspondence information connects each display point $D$ to its corresponding camera chip point $C$.  Depending on the displayed pattern, this can be either done in single-shot for a fixed display pattern (e.g., cross-sinusoid) or multi-shot using, e.g., a phase shifted sinusoid pattern (see \cite{wang2023accurate,huang2018review} for details). Eventually, we extract the angle of reflection $\theta$ for each object point as introduced in Sec.\ref{principle}. We emphasize again that we do not equate $\theta$ with the zenith angle of the respective surface normal, as this would assume an orthographic ray model which leads to large errors. As discussed, $\theta$ is ambiguous and we obtain two possible candidates. We find the correct $\theta$ by applying two geometrical constraints (see also Fig. \ref{fig1}(c)): First, the largest possible $\theta$ is defined by the shortest possible working distance of the camera, while the smallest possible $\theta$ is $0^\circ$ (equivalent to object at infinity). Second, the surface normal calculated from the correct $\theta$ always lies in the plane that is spanned by $D$,$S$, and $C$. For the so obtained correct $\theta$ the ray geometry shown in Fig.~\ref{fig1}(c) for one surface point can then be expressed as

\begin{equation}\label{eq:theta}
    arccos\frac{\overrightarrow{CS}\cdot\overrightarrow{DS}}{\lvert \overrightarrow{CS}\rvert \lvert \overrightarrow{DS}\rvert} = 2\theta ~~.
\end{equation}

\noindent We emphasize again that this equation could not be solved for standard PMD, as $S$ and $\theta$ are both unknowns (normal-depth ambiguity). The novelty of our method is in realizing that SfP gives direct access to the angle of reflection $\theta$ via the DoP given by \eqref{eq:dop} or \eqref{eq:dop_metal}, and thus that PMD and SfP compliment each other in a manner never fully realized in previous work.

This means that, by inserting the measured $\theta$ in \eqref{eq:theta}, the surface normal map and the surface shape can be calculated simultaneously free from any ambiguities and without the need for surface integration methods. Moreover, this procedure does not assume the orthographic projection model and can be done for common perspective cameras. \\

\noindent \textbf{Experiments and Results:} Our prototype consists of a polarization camera (FLIR BFS-U3-51S5PC-C) and an e-ink tablet with unpolarized display (BOOX Tablet Tab X) shown in Fig.~\ref{results}(a).  To quantitatively evaluate our proposed concept, we measured a reflective metal bearing ball with known size (diameter: \( 25.4 mm\)) and known complex refractive index \( m + i \kappa  = 2.76 + 3.79i \). A bearing ball is a precisely manufactured part, meaning that its specified diameter and surface slope can be taken as ground truth for our measurement. The measurement was performed in single-shot: A cross-sinusoidal pattern was displayed to illuminate the object surface (Fig. \ref{results}(a)). Subsequently, we calculated the surface normal map and depth map with our proposed method. Fig. \ref{results} shows (b) one of the four simultaneously captured images of the polarization camera,  (c) the calculated normal map, and (d) the calculated 3D shape. We compared our reconstructed shape and normal map with the corresponding ground truth. The root mean square error (RMSE) of the captured normal map w.r.t. the ground truth normal map was found to be $0.6^\circ$. The reconstructed bearing ball radius is evaluated to $25.47 mm$, while the ground truth is $25.4 mm$ ($70\mu m$ error). The obtained results make our novel method a promising candidate for high-accuracy surface inspection tasks in optical metrology, where our polarization-based procedure can potentially complement or even replace existing (stereo) PMD approaches. 

In addition,  our method also proves to be effective for measuring complex-shaped free-form objects, e.g., for applications in computer vision or graphics. In Fig. \ref{results}(e)-(l), we show the measurement of two complex-shaped specular objects: horse and bird (both chrome-coated: \(  m + i \kappa  = 3.13 + 3.31i\)). The measurements have been performed in a multi-shot fashion by displaying phase-shifted sinusoids. Fig. \ref{results} shows the results. Although we have to restrict ourselves to a qualitative evaluation due to the unknown ground truth, it can be seen that the evaluated normal maps and shapes closely resemble the complex object surfaces.

\section{Summary and Discussion}

We introduced a novel 3D imaging concept for the accurate and fast (single-shot) measurement of complex-shaped specular surfaces. Our method uses polarimetric information from SfP and complements it with geometric information from PMD in a manner never fully realized in previous work.
Our novel approach improves upon state-of-the-art PMD by addressing the height-normal ambiguity problem without the need for a second display or camera. We maintain the single-camera PMD setup without relying on prior knowledge about the object. Simultaneously, our method significantly improves upon state-of-the-art SfP by computing the surface normal and depth simultaneously free from ambiguity and without relying on the orthographic projection model assumption. This significantly improves the error of the estimated normals from up to $25^\circ$ (for standard SfP \cite{atkinson2006recovery}) to below $0.6^\circ$ in our shown measurement.

However, our novel method is not without drawbacks. Compared to standard PMD, we suffer a reduced signal-to-noise ratio (SNR) due to the polarizer grid mounted in front of our camera chip. This leads to longer exposure times or the need for brighter light sources. Compared to standard SfP, our approach requires a calibrated display and does not work with arbitrary unpolarized light sources, such as the sun. Another important evaluation criterion for our method is speed: Although we have demonstrated quantitative measurements of the bearing ball in single-shot using crossed-sinusoidal patterns, multi-shot measurements with phase-shifted sinusoids are generally more robust, especially for complex object shapes with high surface frequencies. In practice, a tradeoff between speed, accuracy, and robustness needs to be found for each respective application. 
In the future, we hope that our novel approach will contribute to further tightening the multidisciplinary bonds between computer vision and optical metrology surface testing, which could potentially lead to further exciting imaging solutions in diverse fields such as in medical imaging, AR/VR, cultural heritage analysis, robotics, or autonomous driving.

\bibliography{ref}


\end{document}